\documentclass{esannV2}
\usepackage[dvips]{graphicx}
\usepackage[latin1]{inputenc}
\usepackage{amssymb,amsmath,array}
\usepackage{subfigure}
\usepackage{array}
\usepackage{multirow}
\usepackage{url}

\begin{document}
%style file for ESANN manuscripts
\title{Modularity-Based Clustering for Network-Constrained Trajectories}

%***********************************************************************
% AUTHORS INFORMATION AREA
%***********************************************************************
\author{Mohamed K. El Mahrsi$^1$ and Fabrice Rossi$^2$
%
% Optional short acknowledgment: remove next line if non-needed
%\thanks{This is an optional funding source acknowledgement.}
%
% DO NOT MODIFY THE FOLLOWING '\vspace' ARGUMENT
\vspace{.3cm}\\
%
% Addresses and institutions (remove "1- " in case of a single institution)
1- T\'el\'ecom ParisTech - D\'epartement Informatique et R\'eseaux \\
46, rue Barrault 75013 Paris - France
%
% Remove the next three lines in case of a single institution
\vspace{.1cm}\\
2- Universit\'e Paris I - D\'epartement SAMM\\
%2- Universit\'e Paris I - D\'epartement Statistique, Analyse, Mod\'elisation Multidisciplinaire\\
90, rue de Tolbiac 75634 Paris CEDEX 13 - France\\
}
%***********************************************************************
% END OF AUTHORS INFORMATION AREA
%***********************************************************************

\maketitle

\begin{abstract}
We present a novel clustering approach for moving object trajectories that are constrained by an underlying road network. The approach builds a similarity graph based on these trajectories then uses modularity-optimization hiearchical graph clustering to regroup trajectories with similar profiles. Our experimental study shows the superiority of the proposed approach over classic hierarchical clustering and gives a brief insight to visualization of the clustering results. 
\end{abstract}

\section{Introduction}

Traffic congestion has become a major problem affecting many human activities on a daily basis and resulting in both serious transportation delays and environmental damages. Continuously collecting information about the state of the road network (e.g. occupancy rates of the road segments) can be valuable for a better understanding of the traffic flow and a wiser planning and restructuring of the road network. Due to the high deployment and maintenance costs of dedicated traffic sensors, a more attractive approach to achieve this aim is to collect traces directly from GPS-equipped vehicles. The collected data can be map matched to corresponding road segments and can be used to deduce the state of the network in real time. It can also be stored and used later to conduct further, more complex data analysis tasks.
.

The problem addressed throughout this article is how to discover  clusters of network-constrained trajectories, i.e. how to group together trajectories that moved along the same parts of the road network. This post-analysis step is conducted on a considerable amount of collected data. It can lead to a better understanding of global movement patterns and tendencies that go unnoticed on the individual level as well as a better grasp of the reasons that lead to congestion situations.

Section \ref{sec:ProblemStatement} presents our formulation of the network-constrained moving object trajectories clustering problem. Our approach to solve this problem is presented in Section \ref{sec:ClusteringApproach}. Experimental results are exposed in Section \ref{sec:Results} whereas related work is briefly discussed in Section \ref{sec:RelatedWork}. Finally, Section \ref{sec:Conclusion} concludes this paper.

\section{Data Model and Problem Statement}
\label{sec:ProblemStatement}
A road network can be represented as a directed graph $G = (V, E)$. $V$ is the set of nodes (or vertices) representing the road intersections in the network whereas $E$ is the set of edges denoting road segments that interconnect these intersections. The direction of a given edge $e=(v_1,v_2)$ linking two vertices $v_1$ and $v_2$ indicates that the corresponding road segment can be travelled from $v_1$ to $v_2$ and not the other way around. A trajectory $T$ traveling along this road network can be modeled as the ordered sequence of visited segments. If travel times are to be taken into account, each segment $e_i$ can be timestamped with the date $t_i$ the trajectory $T$ visited it: $T = \left\{ (t_1, e_1), ... , (t_i, e_i), ... , (t_n, e_n) \right\}$ ($n$ being the number of segments contained in $T$).

Given a dataset of trajectories $\mathcal{T}$ that travelled along a road network $G$, the network-constrained trajectory clustering problem consists in discovering sub-groups (or clusters) $\mathcal{C} = \{C_1, C_2, ... , C_m\}$ of trajectories exhibiting similar behavior. Resemblance between trajectories of the same cluster $C_i$ should be as high as possible and trajectories across two different clusters $C_i$ and $C_j$ should be as different as possible.

\section{Clustering Approach}
\label{sec:ClusteringApproach}

Our clustering approach proceeds in two steps. First, it calculates a similarity graph from $\mathcal{T}$. Then, in the second step, the graph is used to conduct modularity-based graph clustering and regroup similar trajectories together.

\subsection{Similarity between trajectories}
\label{sec:Similarity}

We consider trajectories as bags-of-segments: comparisons between trajectories are done on a segment-basis (i.e. each segment is checked individually, without taking account of the order or the presence of other segments). This choice comes from the fact that: i. in a context of traffic analysis, congestion situations appear first on the level of individual, isolated segments then spread naturally among adjacent road segments. Inspecting each segment apart is, therefore, sufficient to detect these situations; and ii. even if the approach bag-of does not directly take into account the order of the segments, thanks to the fact that the underlying network is a directed graph the order of travel through the segments is implicitly respected.

To assess their relevance to each trajectory, we assign weights to road segments based on their frequency in the data set in a TF-IDF (\textit{Term Frequency - Inverse Document Frequency}) fashion: the spatial weight of a segment $e$ in a trajectory $T$ is defined, analogously to the TF-IDF weight, as follows:
$$
\omega_{e,T} = \frac{\mbox{length}(e)}{\sum_{e' \in T}\mbox{length}(e')} \cdot \log\frac{| \mathcal{T} |}{| \{T' : e \in T' \} |}
$$
$\mbox{length}(e)$ is the spatial length of the segment $e$, $|\mathcal{T}|$ is the cardinality of the dataset $\mathcal{T}$ and $| \{T' : e \in T' \} |$ the cardinality of the subset of trajectories containing the segment $e$. The first term of the multiplication is the equivalent to the term frequency whereas the second is the equivalent to the inverse document frequency.

To compare two trajectories $T_i$ and $T_j$ we calculate their cosine similarity:
$$
\mbox{Similarity}(T_i,T_j) = \frac{\sum_{e \in E} \omega_{e,T_i} \cdot \omega_{e,T_j}}{\sqrt{\sum_{e \in E} \omega_{e,T_i}^2} \cdot \sqrt{\sum_{e \in E} \omega_{e,T_j}^2}}
$$

This weighting and similarity calculation approach takes only the spatial dimension into consideration. The reason is that, for traffic monitoring and optimization purposes, decision making might involve rerouting traffic from a portion of the network to another. Thus affecting all trajectories that passed by the concerned zone whether they travelled together or at different dates.

\subsection{Clustering Algorithm}
As mentioned before, the first step of our clustering algorithm consists in constructing a weighted, undirected similarity graph $G_{\mbox{Sim}} = (\mathcal{T}, E', W)$. Each trajectory from the dataset $\mathcal{T}$ corresponds to a node in $G_{\mbox{Sim}}$. An edge $e \in E'$ links two trajectories $T_i$ and $T_j$ if and only if $\mbox{Similarity}(T_i,T_j) > 0$ in which case the similarity is assigned as a weight ($\omega_e \in W$) to the edge. The choice of graph representation is not only natural but it also puts extra emphasis on the fact that two trajectories that have nothing in common should never be put directly into the same cluster (since no edge provides a direct link between the two).

Since we are interested in analyzing an important number of trajectories and since an edge links two trajectories together if they share at least one road segment in common, $G_{\mbox{Sim}}$ tends to be very large and its vertices have generally high degrees. For these reasons, modularity-based community detection algorithms are  efficient for clustering such graphs \cite{Fortunato_2010}. Modularity measures the classification quality by inspecting the arrangement of edges within clusters (commonly called communities) of vertices. A high modularity indicates that the edges within communities are more numbered (or have more important weights) than in the case of randomly distributed edges.

We opted for an implementation of the modularity-based hierarchical graph clustering algorithm proposed in \cite{Noack_2009} for our clustering step. The algorithm performs modularity optimization on the nodes of the similarity graph and the structure of the discovered communities is validated by means of comparison against random graphs generated with the same set of nodes. If validated, the communities are retained and the algorithm proceeds recursively on each isolated community (i.e. considering only the sub-graph including the nodes of the community). The final result is a hierarchy of nested clusters that can either be explored level by level or using a greedy approach that expands, at each given step, the cluster that induces the minimal loss of modularity.

\section{Experimental Results}
\label{sec:Results}

For our experimental study, we used a synthetic dataset\footnote{The dataset is available at: \url{http://perso.telecom-paristech.fr/~mahrsi/ESANN2012/}} of 10000 trajectories generated with the Brinkhoff generator \cite{Brinkhoff_2002} using the Oldenburg map which contains 6105 nodes and 7035 undirected edges that can be travelled in both directions.

\begin{figure}[h]
\centering
	\subfigure{
		\includegraphics[scale=0.37]{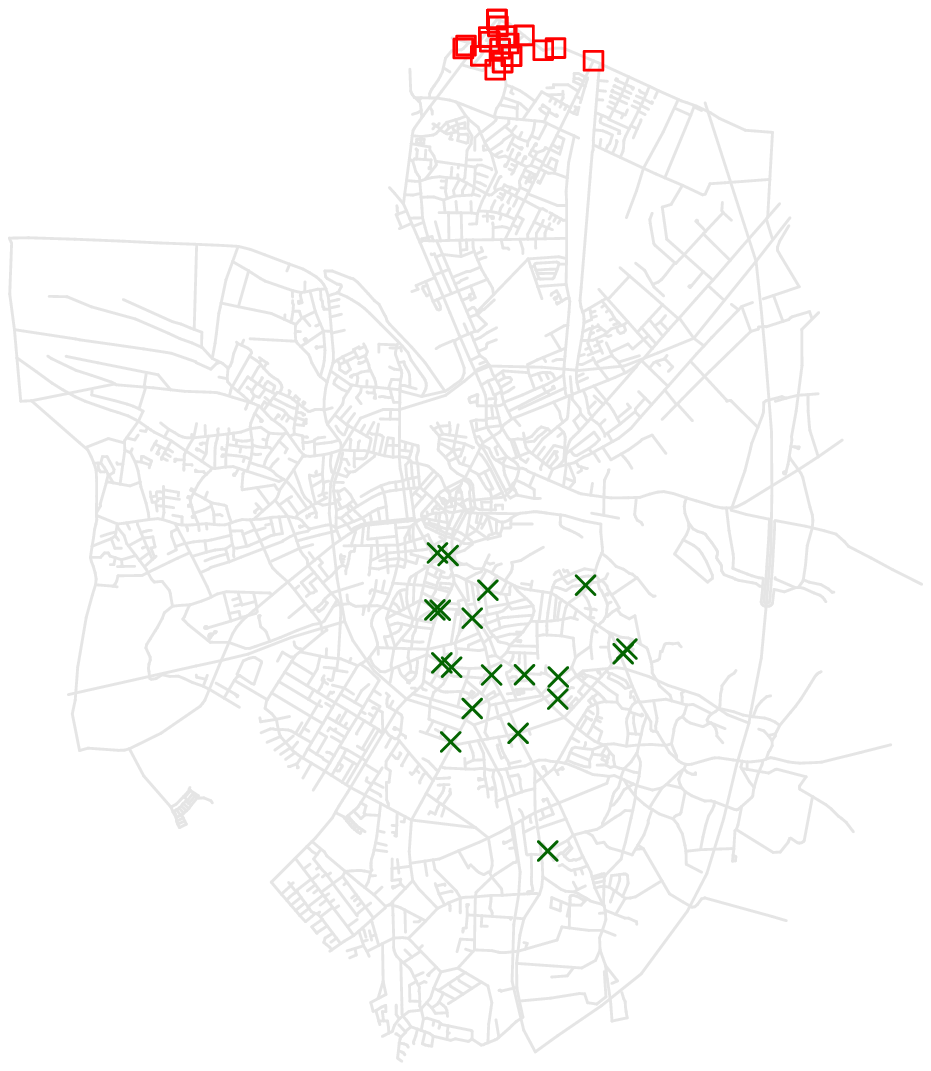}
	}
	\subfigure{
		\includegraphics[scale=0.37]{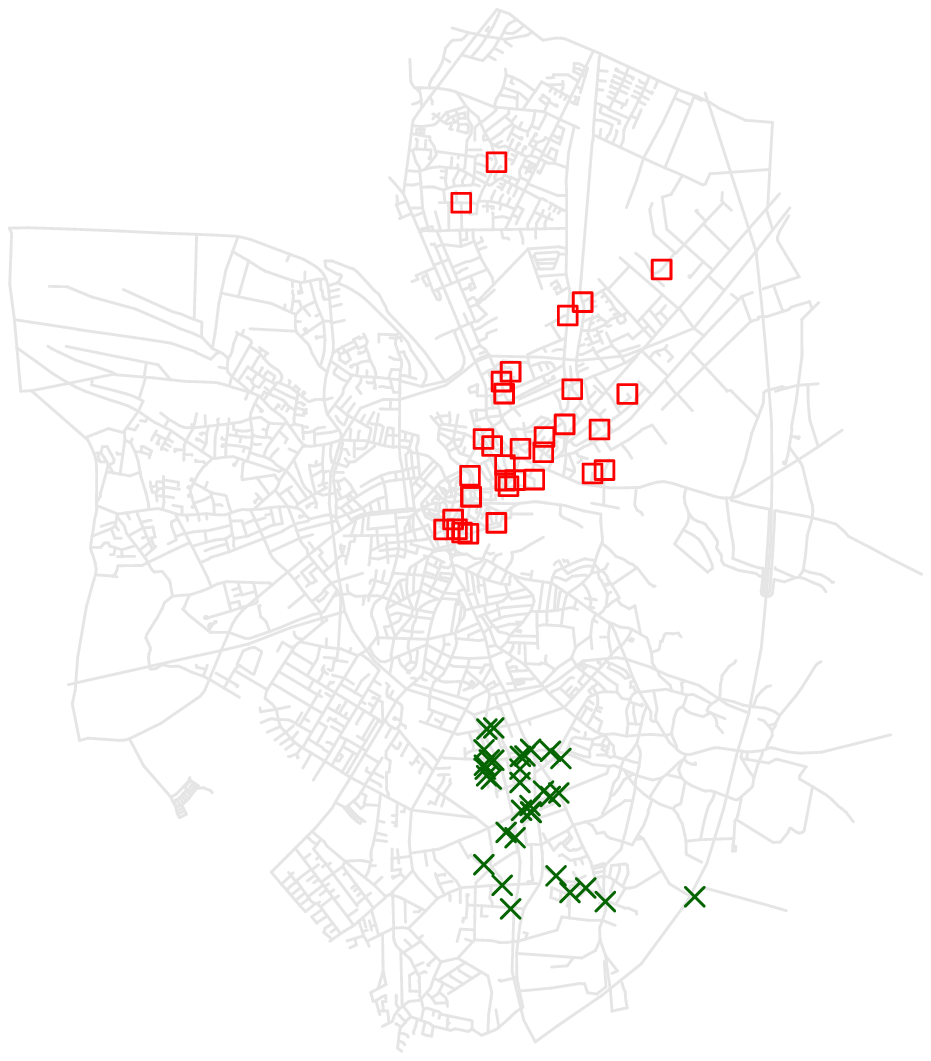}
	}
	\subfigure{
		\includegraphics[scale=0.37]{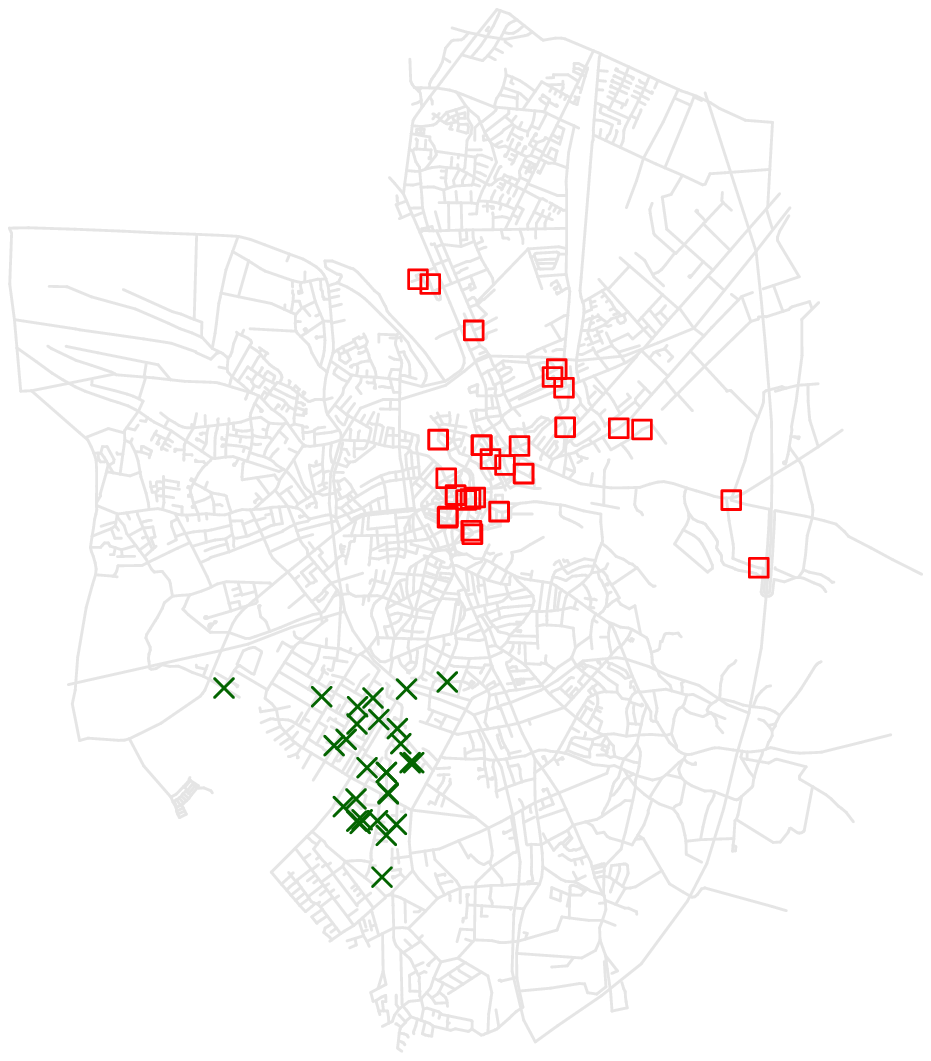}
	}
\caption{Departures points (crosses) and arrival points (empty squares) of some clusters that travelled along the most occupied road segment.}
\label{fig:clusters}
\end{figure}

We started by comparing different similarity calculation approaches: we compared our cosine similarity using spatial weighting (cf. Section \ref{sec:Similarity}) with the Jaccard index  and cosine similarity with classic TD-IDF weighting. The clustering step achieved the highest modularity optimization with spatial weighting (0.5635 vs 0.5251 for Jaccard index and only 0.4861 for classic TF-IDF). The algorithm produced a hierarchy of clusters that spans on 6 levels with 9 clusters on the top level and 648 clusters in the lowest level. Figure \ref{fig:clusters} shows examples of clusters produced by our algorithm on the third level of the clustering hierarchy (we chose to expand level by level since this approach seemed to give the most balanced clusters for this dataset). Each sub-figure shows the distribution of departure and arrival points of the trajectories, in a given cluster, that moved along the most visited road segment.

Visualization of the generated clusters shows promising results as of the capability of the clustering approach to divide trajectories into well separated groups. Moreover, the fact that the algorithm produces a hierarchy of nested clusters can be very useful for the understanding and visualization of traffic: one can start with a given level and locate interesting clusters (e.g. clusters that crossed a given area of interest) and see how these clusters are expanded into sub-clusters on lower levels.

We also compared our clustering approach to classic hierarchical agglomerative clustering. To this end, we used a distance that is complementary to our similarity measure: $ \mbox{distance}(T_1,T_2) = 1-\mbox{Similarity}(T_1,T_2)$. We calculated an adjacency matrix based on this distance and we used it for the agglomerative clustering (with single, average and full linkage). Comparisons were conducted for the same number of cluster: for each of the 6 hierarchy levels produced in our approach, we cut the hierarchical clustering at the same number of clusters and we compared the two. Quality measures that we used are: i. interclass and intraclass inertia of the start points of the trajectories; ii. interclass and intraclass inertia of the end points of the trajectories; and 3. interclass and intraclass overlap of the trajectories. The first two are used to assess the compactness of the regrouped trajectories start/end points . Overlap measures give an appreciation of how much road segments trajectories share among and across the resulting clusters ($\mathcal{C}$ being the set of resulting clusters, and $|C|$ the number of trajectories in the cluster $C$):
$$
\mbox{intraclass ovelap} = \sum_{C \in \mathcal{C}} \frac{1}{|C|} \sum_{T_i,T_j \in C} \frac{\sum_{e \in T_i, e \in T_j}\mbox{length}(e)}{\sum_{e \in T_i}\mbox{length}(e)}
$$
$$
\mbox{interclass overlap} = \sum_{C_i \in \mathcal{C}} \frac{1}{|\mathcal{T}| - |C_i|} \sum_{C_j \in \mathcal{C}, j \neq i} \sum_{T \in C_i, T' \in C_j}\frac{\sum_{e \in T, e \in T'}\mbox{length}(e)}{\sum_{e \in T}\mbox{length}(e)}
$$

Due to lack of space, we only show the results of our interclass and intraclass overlap comparison (Table \ref{tab:overlap}). Modularity-optimization clustering achieves the best intraclass overlap among the tested approaches. This indicates that trajectories within a same cluster are more similar and share more road segments than in the case of classic hierarchical clustering. The lower (at first glance better) interclass overlap achieved by single linkage hierarchical clustering comes from the fact that this approach produces very unbalanced clusters (it tends to produce a huge cluster regrouping most of the trajectories in the dataset while the other clusters are very small). This problem is also visible with average and full linkage hierarchical clustering but only when the number of clusters is small. When the number of clusters gets bigger, average and full linkage as well as modularity-optimization clustering all behave in a similar manner w.r.t. interclass overlap.

\begin{table}[htdp]
\begin{center}
\small
\begin{tabular}{|c|c|c|c|c|c|c|c|c|}
\hline
Nbr. of & \multicolumn{4}{c|}{Intraclass overlap} &  \multicolumn{4}{c|}{Interclass overlap}\\
\cline{2-9}
 clusters & HC(S) & HC(A) & HC(F) & Mod. & HC(S) & HC(A) & HC(F) & Mod.\\
\hline
9 & 111 & 111 & 123 & 608 & 0.5 & 0.2 & 38.9 & 42.8\\
\hline
45 & 149 & 155 & 349 & 1768 & 5.1 & 2.9 & 78.4 & 62.2\\
\hline
159 & 270 & 1594 & 1877 & 3121 & 8.8 & 59.1 & 97.9 &79.4\\
\hline
419 & 569 & 3073 & 3682 & 4264 & 22.4 & 69.5 &  87.1 & 87.9\\
\hline
621 & 822 & 3999 & 4419 & 4780 & 30.3 & 76.2 & 85.8 & 89.7\\
\hline
648 & 881 & 4106 & 4491 & 4823 & 35.4 & 76.5 & 85.8 & 89.8\\
\hline
\end{tabular}
\end{center}
\caption{Interclass and intraclass overlaps achieved by modularity-optimization trajectory clustering and hierarchical agglomerative clustering (with S: single linkage, A: average linkage and F: full linkage).}
\label{tab:overlap}
\end{table}

\section{Related Work}
\label{sec:RelatedWork}
Clustering trajectory data attracted many research in the last few years. Existing proposals include TraClus \cite{Lee_2007}, convoy \cite{Jeung_2008b} and flock \cite{Benkert_2008} patterns and many others. These are mainly density-based approaches that suppose that the moving objects can move freely on an euclidean space. Therefore, these approaches are substantially different from the problem at hand where the movement is constrained. The case of constrained trajectories started to attract attention only recently. In \cite{Kharrat_2008}, the authors propose a density-based approach to discover dense paths in such trajectories. Like all density-based method, the approach is very sensitive to the configuration of the $minPts$ and $\epsilon$ parameters. Furthermore, the approach clusters sub-trajectories together and does not conserve trajectory participation on the whole dense path (i.e. a trajectory might participate only partially in the path). Our approach, on the contrary, regroups whole trajectories. To our knowledge, our work is the first to apply modularity-optimization graph clustering in the context of trajectory data.

\section{Conclusion}
\label{sec:Conclusion}

In this article, we presented a novel approach to cluster trajectories constrained by an underlying road network. The approach starts by computing a similarity graph between trajectories based on the cosine spatial similarity that we defined. The graph is then used to conduct hierarchical modularity-optimization clustering to discover communities of trajectories that exhibited similar behavior. Results on synthetic trajectory data are promising and showed that the proposed approach yields better, more relevant clusters than the classic hierarchical clustering. For future research directions, we are mainly interested in the visual exploitation of the clustering results as well as the study of their relevance in decision making in a context of traffic rerouting and optimization.

% ****************************************************************************
% BIBLIOGRAPHY AREA
% ****************************************************************************

\begin{footnotesize}

% IF YOU DO NOT USE BIBTEX, USE THE FOLLOWING SAMPLE SCHEME FOR THE REFERENCES
% ----------------------------------------------------------------------------

% ----------------------------------------------------------------------------

% IF YOU USE BIBTEX,
% - DELETE THE TEXT BETWEEN THE TWO ABOVE DASHED LINES
% - UNCOMMENT THE NEXT TWO LINES AND REPLACE 'Name_Of_Your_BibFile'

\bibliographystyle{unsrt}
\bibliography{bibliography}

\begin{thebibliography}{1}

\bibitem{Fortunato_2010}
Santo Fortunato.
\newblock Community detection in graphs.
\newblock {\em Physics Reports}, 486(3-5):75--174, 2010.

\bibitem{Noack_2009}
Andreas Noack and Randolf Rotta.
\newblock Multi-level algorithms for modularity clustering.
\newblock In {\em Proceedings of the 8th International Symposium on
  Experimental Algorithms}, SEA '09, pages 257--268, Berlin, Heidelberg, 2009.
  Springer-Verlag.

\bibitem{Brinkhoff_2002}
Thomas Brinkhoff.
\newblock A framework for generating network-based moving objects.
\newblock {\em Geoinformatica}, 6:153--180, June 2002.

\bibitem{Lee_2007}
Jae-Gil Lee, Jiawei Han, and Kyu-Young Whang.
\newblock Trajectory clustering: a partition-and-group framework.
\newblock In {\em SIGMOD '07: Proceedings of the 2007 ACM SIGMOD international
  conference on Management of data}, pages 593--604, New York, NY, USA, 2007.
  ACM.

\bibitem{Jeung_2008b}
Hoyoung Jeung, Man~Lung Yiu, Xiaofang Zhou, Christian~S. Jensen, and Heng~Tao
  Shen.
\newblock Discovery of convoys in trajectory databases.
\newblock {\em Proc. VLDB Endow.}, 1(1):1068--1080, 2008.

\bibitem{Benkert_2008}
Marc Benkert, Joachim Gudmundsson, Florian H\"{u}bner, and Thomas Wolle.
\newblock Reporting flock patterns.
\newblock {\em Comput. Geom. Theory Appl.}, 41(3):111--125, 2008.

\bibitem{Kharrat_2008}
Ahmed Kharrat, Iulian~Sandu Popa, Karine Zeitouni, and Sami Faiz.
\newblock Clustering algorithm for network constraint trajectories.
\newblock In Anne Ruas and Christopher~M. Gold, editors, {\em SDH}, Lecture
  Notes in Geoinformation and Cartography, pages 631--647. Springer, 2008.

\end{thebibliography}

\end{footnotesize}

% ****************************************************************************
% END OF BIBLIOGRAPHY AREA
% ****************************************************************************

\end{document}